\def\year{2020}\relax
\documentclass[letterpaper]{article} 
\usepackage{aaai20}  
\usepackage{times}  
\usepackage{helvet} 
\usepackage{courier}  
\usepackage[hyphens]{url}  
\usepackage{graphicx} 
\def\UrlFont{\rm}  
\usepackage{graphicx}  
\usepackage{xspace}
\usepackage{wrapfig}
\nocopyright

\usepackage{subfiles}
\usepackage{subfig}
\usepackage{amsmath}
\usepackage{amsfonts}
\usepackage{multirow,booktabs,threeparttable}
\usepackage{gensymb}
\usepackage{siunitx}
\newcommand{\eg}{\textit{e}.\textit{g}. }
\def\etal{\emph{et al}.}
\usepackage{xcolor}
\usepackage{soul}
\newcommand\model{SATRN\xspace}

\title{On Recognizing Texts of Arbitrary Shapes with 2D Self-Attention}

\author{\textbf{Junyeop Lee, Sungrae Park, Jeonghun Baek, Seong Joon Oh$^{\dagger}$, Seonghyeon Kim, Hwalsuk Lee} 
\\ 
Clova AI Research, NAVER/LINE Corp\\ 
\{junyeop.lee, sungrae.park, jh.baek, seonghyeon.kim, hwalsuk.lee\}@navercorp.com, 
$^{\dagger}$coallaoh@linecorp.com
}

\begin{document}

\maketitle

\begin{abstract}

Scene text recognition (STR) is the task of recognizing character sequences in natural scenes. 
While there have been great advances in STR methods, current methods still fail to recognize texts in arbitrary shapes, such as heavily curved or rotated texts, which are abundant in daily life (\eg restaurant signs, product labels, company logos, etc).
This paper introduces a novel architecture to recognizing texts of arbitrary shapes, named \emph{Self-Attention Text Recognition Network (\model)}, which is inspired by the Transformer.
\model utilizes the self-attention mechanism to describe two-dimensional (2D) spatial dependencies of characters in a scene text image. 
Exploiting the full-graph propagation of self-attention, \model can recognize texts with arbitrary arrangements and large inter-character spacing. 
As a result, 
\model outperforms existing STR models by a large margin of 5.7 pp on average in ``irregular text'' benchmarks.
We provide empirical analyses that illustrate the inner mechanisms and the extent to which the model is applicable (\eg rotated and multi-line text). We will open-source the code.

\end{abstract}

\section{Introduction}

Scene text recognition (STR) addresses the following problem: given an image patch tightly containing text taken from natural scenes (\eg license plates and posters on the street), what is the sequence of characters?~\cite{zhu2016scene,long2018scene}
Applications of deep neural networks have led to great improvements in the performance of STR models~\cite{RARE,R2AM,ATR,FAN,Char-Net,EP}. They typically combine a convolutional neural network (CNN) feature extractor, designed for abstracting the input patch, with a subsequent recurrent neural network (RNN) character sequence generator, responsible for character decoding and language modeling. The model is trained in an end-to-end manner.

While these methods have brought advances in the field, they are built upon the assumption that input texts are written horizontally. Cheng \etal~\cite{FAN} and Shi \etal~\cite{RARE,Aster}, for
example, have collapsed the height component of the 2D CNN feature maps into a 1D feature map. They are conceptually and empirically inept at interpreting texts with arbitrary shapes, which are important challenges in realistic deployment scenarios. 

\begin{figure}
\includegraphics[width=1.0\linewidth]{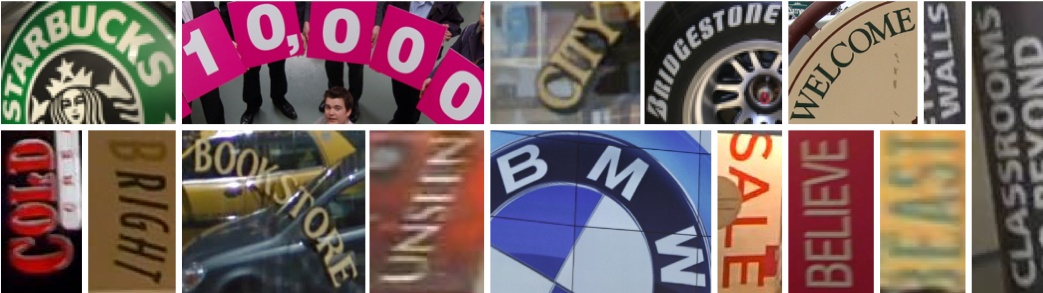}
\caption{\small Texts of arbitrary shapes: remaining challenges for scene text recognition. 
}
\label{tmp_figure}
\vspace{-1em}
    \centering
    \subfloat[\small SAR (Li et al., 2019)] {\includegraphics[width=1.0\linewidth]{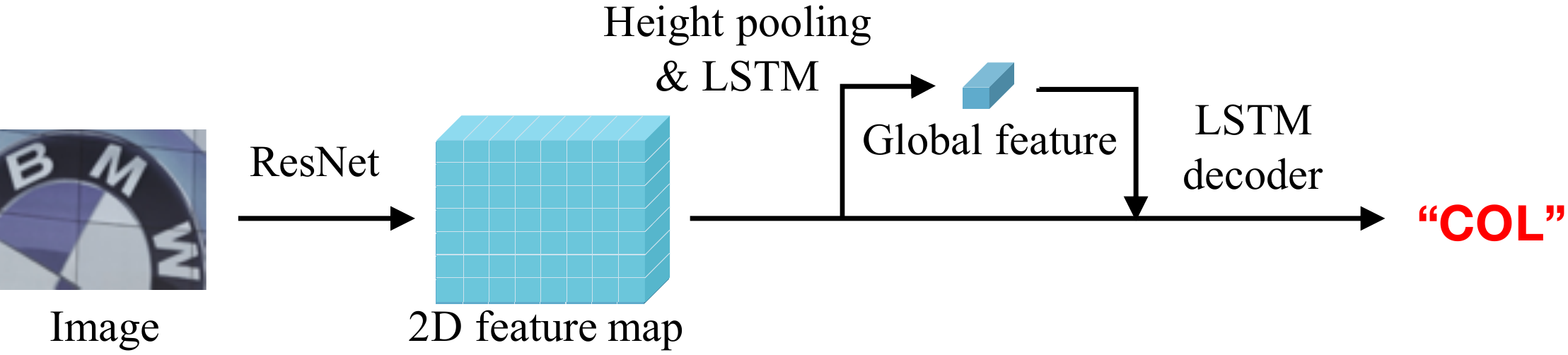}} \\
    \vspace{-0.5em}
    \subfloat[\small \model (Ours)] {\includegraphics[width=1.0\linewidth]{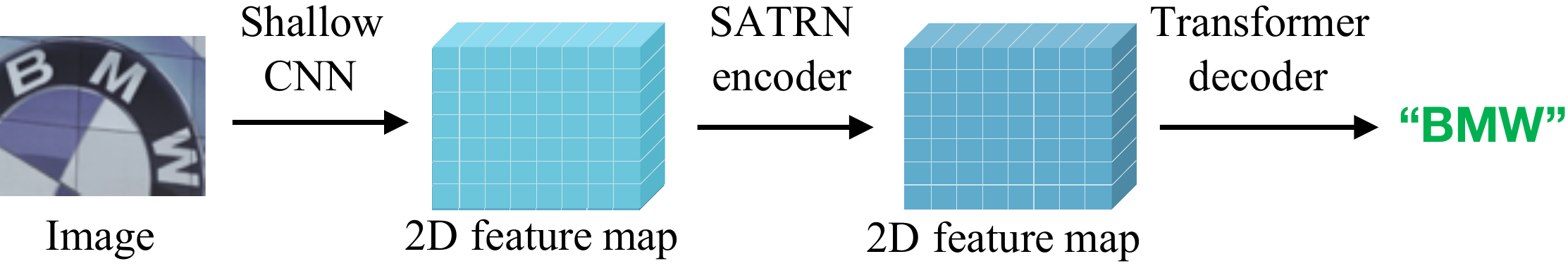}}
    \caption{\small \model addresses the text images of difficult shapes (curved ``BMW'' logo) by adopting a self-attention mechanism, while keeping intermediate feature maps two dimensional. \model thus models long-range dependencies spanning 2D space, a feature necessary for recognizing texts of irregular geometry.
    } 
    \label{fig:simple_archi_cmp}
    \vspace{-2em}
\end{figure}

Realizing the significance and difficulty of recognizing texts of arbitrary shapes, the STR community has put more emphasis on such image types. The introduction of ``irregular shape'' STR benchmarks~\cite{Benchmark_Baek} is an evidence of such interest. 
On the method side, recent STR approaches are focusing more on addressing texts of irregular shapes. There are largely two lines of research: (1) input rectification and (2) usage of 2D feature maps. Input rectification~\cite{RARE,Aster,Char-Net,STAR-Net,RECUR-STN} uses spatial transformer networks (STN,~\cite{STN}) to normalize text images into canonical shapes: horizontally aligned characters of uniform heights and widths. These methods, however, suffer from the limitation that the possible family of transformations have to be specified beforehand. 

Methods using 2D feature maps~\cite{AON,ATR,SAR}, on the other hand, take the original input image without any modification, learn 2D feature maps, and sequentially retrieve characters on the 2D space. While the usage of 2D feature maps certainly increases room for more complex modelling, specific designs of existing methods are still limited by either the assumption that input texts are written horizontally (SAR~\cite{SAR}), overly complicated model structure (AON~\cite{AON}), or requirement of ground truth character bounding boxes (ATR~\cite{ATR}). We believe the community has lacked a simple solution to nicely handle texts of arbitrary shapes. 
In this paper, we propose an STR model that adopts a 2D self-attention mechanism to resolve the remaining challenging case within STR. Our architecture is heavily inspired by the Transformer~\cite{Transformer},
which has made profound advances in the natural language processing~\cite{char-Transformer,BERT} and vision~\cite{Image-Transformer} fields. Our solution,\textit{ Self-Attention Text Recognition Network (\model)}, adopts the encoder-decoder construct of Transformer to address the cross-modality between the image input and the text output. The intermediate feature maps are two dimensional throughout the network. By never collapsing the height dimension, we better preserve the spatial information than prior approaches~\cite{SAR}. Figure~\ref{fig:simple_archi_cmp} describes how \model preserves spatial information throughout the forward pass, unlike prior approaches.

While \model is performant due to the decoder following original character-level Transformer, we have discovered that a few novel modifications on the Transformer encoder is necessary to fully realize the benefit of self-attention in a 2D feature map. Three new modules are introduced: (1) shallow CNN, (2) adaptive 2D positional encoding, and (3) locality-aware feedforward layer. We will explain them in greater detail in the main text.

The resulting model, \model, is architecturally simple, memory efficient, and accurate. We have evaluated \model for its superior accuracy on the seven benchmark datasets and our newly introduced rotated and multi-line texts, along with its edge on computational cost. We justify the design choices in the encoder through ablative experiments. We note that \model is the state of the art model in five out of seven benchmark datasets considered, with notable gain of 5.7 pp average boost on ``irregular'' benchmarks over the prior state of the art.

We contribute (1) \model, inspired by Transformer, to address remaining challenges for STR; (2) novel modules in \model encoder to make Transformer effective and efficient for STR; and (3) experimental analysis on the effect of proposed modules and verification that \model is particularly good at texts of extreme shapes.

\section{Related Works}

In this section, we present prior works on scene text recognition, focusing on how they have attempted to address texts of arbitrary shapes. Then, we discuss previous works on using Transformer on visual tasks and compare how our approach differs from them.

\subsubsection{Scene text recognition on arbitrary shapes}

Early STR models have assumed texts are horizontally aligned. These methods have extracted width-directional 1D features from an input image and have transformed them into sequences of characters~\cite{RARE,R2AM,ATR,FAN,Char-Net,EP,NRTR,Benchmark_Baek}. 
By design, such models fail to address curved or rotated text. To overcome this issue, spatial transformation networks (STN) have been applied to align text image into a canonical shape (horizontal alignment and uniform character widths and heights)~\cite{RARE,Aster,Char-Net,STAR-Net,RECUR-STN}. 
STN does handle non-canonical text shapes to some degree, but is limited by the hand-crafted design of transformation space and the loss in fine details due to image interpolation.

Instead of the input-level normalization, recent works have spread the normalization burden across multiple layers, by retaining two-dimensional feature maps up to certain layers in the network and information propagation across 2D space.
Cheng et al.~\cite{AON} have first computed four 1D features by projecting an intermediate 2D feature map in four directions. They have introduced a selection module to dynamically pick one of the four features. Their method is still confined to those four predefined directions. Yang et al.~\cite{ATR}, on the other hand, have developed a 2D attention model over 2D features. The key disadvantage of their method is the need for expensive character-level supervision.
Li et al.~\cite{SAR} have directly applied attention mechanism on 2D feature maps to generate text. However, their method loses full spatial information due to height pooling and RNN, thus being inherently biased towards horizontally aligned texts. 
These previous works have utilized a sequence generator sequentially attending to certain regions on the 2D feature map following the character order in texts. 
In this work, we propose a simpler solution with the self-attention mechanism~\cite{Transformer} applied on 2D feature maps. 
This approach enables character features to be aware of their spatial order and supports the sequence generator to track the order without any additional supervision.

\subsubsection{Transformer for visual tasks}

Transformer has been introduced in the natural language processing field~\cite{Transformer,BERT,char-Transformer}. By allowing long-range pairwise dependencies through self-attention, it has achieved breakthroughs in numerous benchmarks. The original Transformer is a sequence-to-sequence model consisting of an encoder and decoder pair, without relying on any recurrent module. 

Transformer has been adopted by methods solving general vision tasks such as action recognition~\cite{nonlocal}, object detection~\cite{nonlocal}, semantic segmentation~\cite{nonlocal,AACN}, and image generation~\cite{SAGAN,Image-Transformer}.
Self-attention mechanism has been extended to two dimensional feature maps to capture long-range spatial dependencies. Since naive extension to spatial features induces high computational cost, these works have considered reducing number of pairwise connections through convolution layers~\cite{nonlocal} or pair pruning~\cite{AACN}. We have adopted the techniques to STR task in \model; details will be discussed later.

\section{SATRN Method}

This section describes our scene text recognition (STR) model, \textit{self-attention text recognition network (\model)}, in full detail. Many of the modules and design choices have been inherited and inspired from the successful Transformer model~\cite{Transformer}, but there are several novel modifications for successful adaptation of STR task. We will provide an overview of the \model architecture, and then focus on the newly introduced modules.

\subsection{\model Overview}

\begin{figure}[t!]
    \centering
    \includegraphics[width=1\linewidth]{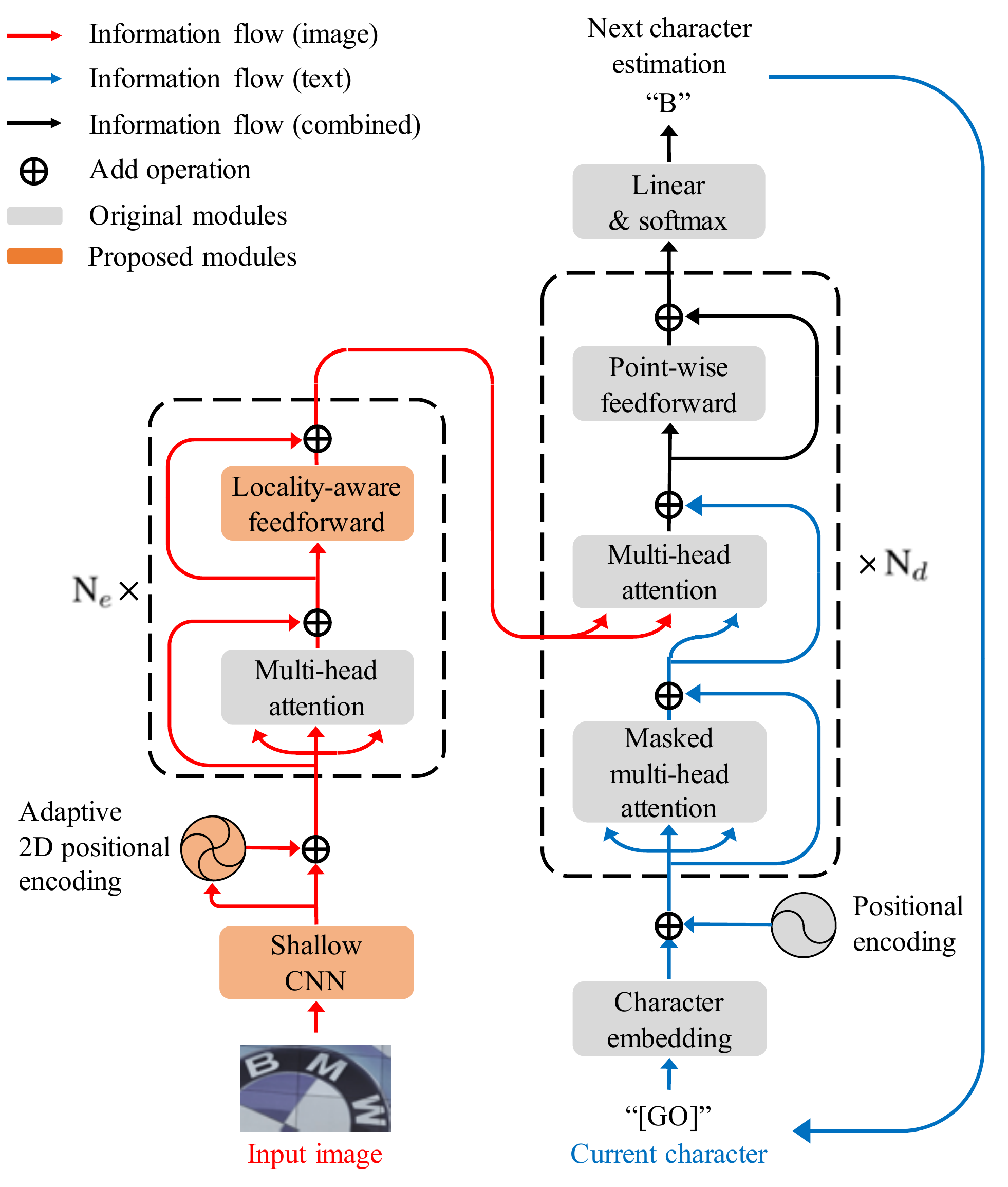}
    \caption{\small \model architecture overview. Left column is encoder and right column is decoder.
    } 
    \label{fig:architecture}
    \vspace{-1em}
\end{figure}

Figure~\ref{fig:architecture} shows the overall architecture of \model. It consists of an encoder (left column), which embeds an image into a 2D feature map, and a decoder (right column), which then extracts a sequence of characters from the feature map.

\subsubsection{Encoder}
The encoder processes input image through a \textit{shallow CNN} that captures local patterns and textures. The feature map is then passed to a stack of self-attention modules, together with an \textit{adaptive 2D positional encoding}, a novel positional encoding methodology developed for STR task. The self-attention modules are modified version of the original Transformer self-attention modules, where the point-wise feed forward is replaced by our \textit{locality-aware feedforward layer}. The self-attention block is repeated $N_e$ times (without sharing weights). In the next section, we will describe in detail the components of \model that are newly introduced in the encoder on top of the original Transformer.

\subsubsection{Decoder}
The decoder retrieves the enriched two-dimensional features from the encoder to generate a sequence of characters. 
The cross-modality between image input and text output happens at the second multi-head attention module. The module retrieves the next character's visual feature 
The feature of the current character is used to retrieve the next character's visual features upon the 2D feature map. 
Most of the decoder modules, such as multi-head attention and point-wise feedforward layers, are identical to the decoder of Transformer~\cite{Transformer}, as the decoder in our case also deals with sequence of characters~\cite{char-Transformer}. Our methodological contributions are focused on adapting the encoder to extract sequential information embedded in images along arbitrary shapes. 

\subsection{Designing Encoder for STR}

We explain how we have designed the encoder to effectively and efficiently extract sequential information from images. There are three main constructs that modify the original Transformer architecture. Each of them will be explained.

\subsubsection{Shallow CNN block} 

Input images are first processed through a shallow CNN. This stage extracts elementary patterns and textures in input images for further processing in the subsequent self-attention blocks. Unlike in natural language processing, visual inputs tend to require much more abstraction as there are many background features to suppress (e.g. background texture of menu plate). Therefore, directly applying the Transformer architecture will put great burden to the expensive self-attention computations. This shallow CNN block performs pooling operations to reduce such burden.

More specifically, the shallow CNN block consists of two convolution layers with 3$\times$3 kernels, each followed by a max pooling layer with 2$\times$2 kernel of stride 2. The resulting 1/4 reduction factor has provided a good balance in computation-performance trade-off in our preliminary studies. If spatial dimensions are further reduced, performance drops heavily; if reduced less, computation burden for later self-attention blocks increases a lot.

\subsubsection{Adaptive 2D positional encoding}
The feature map produced by the shallow CNN is fed to self-attention blocks. The self-attention block, however, is agnostic to spatial arrangements of its input (just like a fully-connected layer). Therefore, the original Transformer has further fed \textit{positional encodings}, an array containing modified index values, to the self-attention module to supply the lacking positional information. 

Positional encoding (PE) has not been essential in vision tasks~\cite{SAGAN,nonlocal,AACN}; the focus in these cases has been to provide long-range dependencies not captured by convolutions. 
On the other hand, positional information plays an important role in recognizing text of arbitrary shape, since the self-attention itself is not supplied the absolute location information: given current character location exactly where in the image can we find the next character? Missing the positional information makes it hard for the model to sequentially track character positions. \model thus employs a 2D extension of the positional encoding.

However, naive application of positional encoding cannot handle the diversity of character arrangements. For example, 10 pixels along width dimension for horizontal text will contain less number of characters than for diagonal text on average. Therefore, different length elements should be used in the positional encoding depending on the type of input.
We thus propose the adaptive 2D positional encoding (A2DPE) to dynamically determine the ratio between height and width element depending on the input. 

We first describe the self-attention module without positional encoding.
We write $\textbf{E}$ for the 2D feature output of shallow CNN and $\textbf{e}_{hw}$ for its entry at position $(h, w)\in [1,...,H]\times [1,...,W]$. The self-attention is computed as
\begin{equation}
{\textbf{att-out}}_{hw}=\sum_{h'w'} \text{softmax} ({\text{rel}}_{(h'w')\rightarrow (hw)}) {\textbf{v}}_{h'w'},
\end{equation}
where the value array $\textbf{v}_{hw}=\textbf{e}_{hw} \textbf{W}^{\text{v}}$ is a transformation of the input feature through linear weights $\textbf{W}^{\text{v}}$ and ${\text{rel}}_{(h'w')\rightarrow (hw)}$ is defined as 
\begin{equation}
{\text{rel}}_{(h'w')\rightarrow (hw)} \propto \textbf{e}_{hw} \textbf{W}^{\text{q}} {\textbf{W}^{\text{k}}}^{\text{T}} {\textbf{e}_{h'w'}}^{\text{T}},
\end{equation}
where $\textbf{W}^{\text{q}}$ and $\textbf{W}^{\text{k}}$ are linear weights that map the input into queries $\textbf{q}_{hw}=\textbf{e}_{hw} \textbf{W}_q$ and keys $\textbf{k}_{hw}=\textbf{e}_{hw} \textbf{W}_k$. Intuitively, ${\text{rel}}_{(h'w')\rightarrow (hw)}$ dictates how much feature at $(h',w')$ attends to feature at $(h,w)$. 

We now introduce our positional encoding A2DPE $\textbf{p}_{hw}$ in this framework as below:
\begin{equation}
{\text{rel}}_{(h'w')\rightarrow (hw)} \propto (\textbf{e}_{hw} + \textbf{p}_{hw} ) \textbf{W}^{\text{q}} {\textbf{W}^{\text{k}}}^{\text{T}} {(\textbf{e}_{h'w'} + \textbf{p}_{h'w'} )}^{\text{T}}.
\end{equation}
Note that A2DPE are added on top of the input features. Now, A2DPE itself is defined as
$\boldsymbol{\alpha}$ and $\boldsymbol{\beta}$.
\begin{equation}
    {\textbf{p}}_{hw} =
    \boldsymbol{\alpha}(\textbf{E}){\textbf{p}}_{h}^{\text{sinu}} + \boldsymbol{\beta}(\textbf{E}){\textbf{p}}_{w}^{\text{sinu}},
\end{equation}
where ${\textbf{p}}_{h}^{\text{sinu}}$ and ${\textbf{p}}_{w}^{\text{sinu}}$ are sinusoidal positional encoding over height and width, respectively, as defined in~\cite{Transformer}.
\begin{gather}
{\textbf{p}}^{\text{sinu}}_{p, 2i} = \text{sin} (p / 10000 ^ {2i / D}), \\
{\textbf{p}}^{\text{sinu}}_{p, 2i + 1} = \text{cos} (p / 10000 ^ {2i / D}),
\end{gather}
where $p$ and $i$ are indices along position and hidden dimensions, respectively. 
The scale factors, $\boldsymbol{\alpha}(\textbf{E})$ and $\boldsymbol{\beta}(\textbf{E})$, are computed from the input feature map $\textbf{E}$ with 2-layer perceptron applied on global average pooled input feature as the followings: 
\begin{gather}
\boldsymbol{\alpha}(\textbf{E}) = \text{sigmoid}\left(\text{max}(0, \boldsymbol{g}(\textbf{E}) \textbf{W}_{\text{1}}^{\text{h}})\textbf{W}_{\text{2}}^{\text{h}}\right), \\
\boldsymbol{\beta}(\textbf{E}) = \text{sigmoid}\left(\text{max}(0, \boldsymbol{g}(\textbf{E}) \textbf{W}_{\text{1}}^{\text{w}})\textbf{W}_{\text{2}}^{\text{w}}\right),
\end{gather}
where $\textbf{W}_{\text{1}}^{\text{h}}$, $\textbf{W}_{\text{2}}^{\text{h}}$, $\textbf{W}_{\text{1}}^{\text{w}}$ and $\textbf{W}_{\text{2}}^{\text{w}}$ are linear weights. The $g(\textbf{E})$ indicates an average pooling over all features in $\textbf{E}$. 
The outputs go through a sigmoid operation.
The identified $\boldsymbol{\alpha}$ and $\boldsymbol{\beta}$ affects the height and width positional encoding directly to control the relative ratio between horizontal and vertical axes to express the spatial diversity. By learning to infer $\boldsymbol{\alpha}$ and $\boldsymbol{\beta}$ from the input, A2DPE allows the model to adapt the length elements along height and width directions.

\subsubsection{Locality-aware feedforward layer}

\begin{figure}
    \centering
    \subfloat[\small Fully-connected \label{fig:ffn}] {\includegraphics[width=0.32\linewidth]{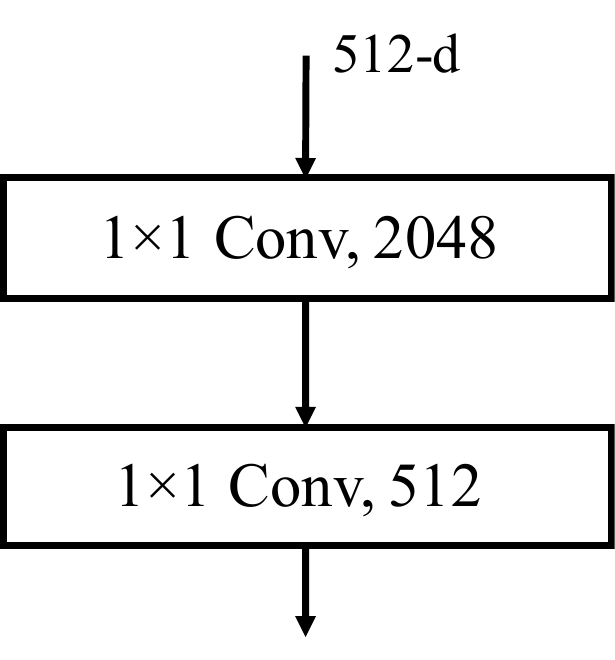}} \hfill
    \centering
    \subfloat[\small Convolution \label{fig:conv}] {\includegraphics[width=0.32\linewidth]{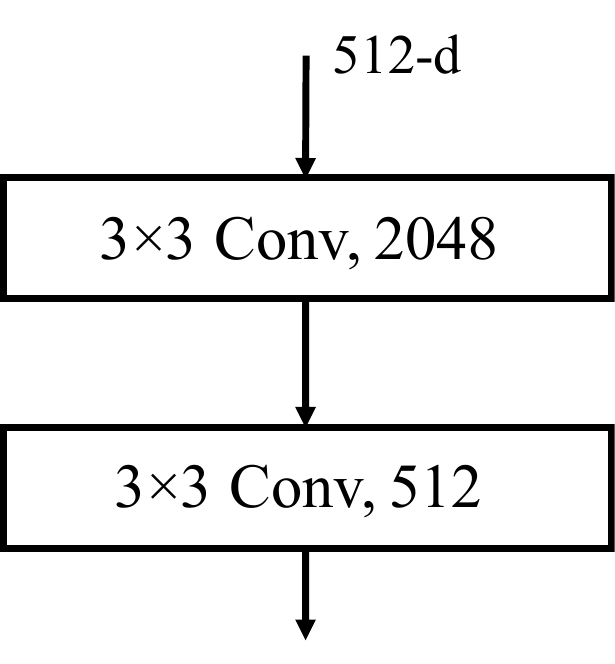}} \hfill
    \centering
    \subfloat[\small Separable \label{fig:ffl}] {\includegraphics[width=0.32\linewidth]{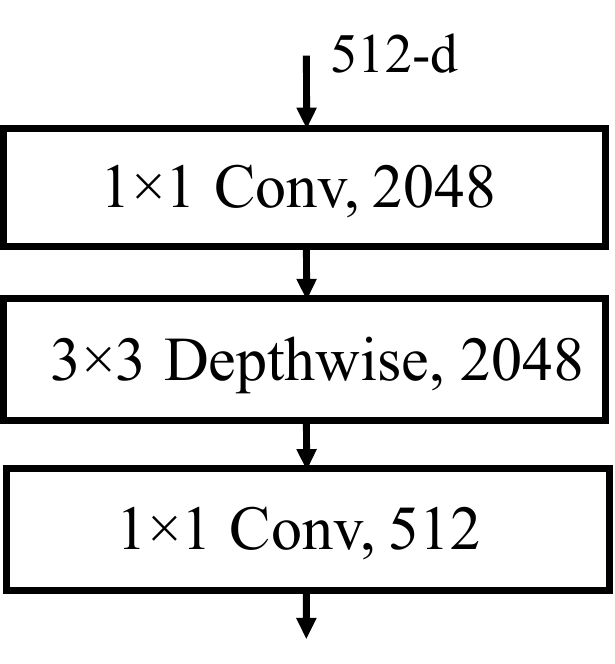}}
    \caption{\small Feedforward architecture options applied after the self-attention layer.}
    \label{fig:ff}
    \vspace{-1em}
\end{figure}

For good STR performance, a model should not only utilize long-range dependencies but also local vicinity around single characters. Self-attention layer itself is good at modelling long-term dependencies, but is not equipped to give sufficient focus on local structures. We have thus improved the original point-wise feedforward layer (Figure~\ref{fig:ffn}), consisting of two $1\times 1$ convolutional layers, by utilizing $3\times 3$ convolutions (Figures~\ref{fig:conv},~\ref{fig:ffl}). In the experiments, we will show that between the naive $3\times 3$ convolution and the depth-wise variant, the latter gives a better performance-efficiency trade-off.

\section{Experiments}

 \begin{table*}[h]
    \tabcolsep=0.3cm
	\newcommand{\tabincell}[2]{\begin{tabular}{@{}#1@{}}#2\end{tabular}}
	\begin{center}
		\small
		\begin{tabular}{l|c|c|cccc|ccc}
			\hline
			 \multirow{2}{*}{\textbf{Method}} & \textbf{Feature} & \textbf{Training} & \multicolumn{4}{c|}{\textbf{Regular test dataset}} &  \multicolumn{3}{c} {\textbf{Irregular test dataset}}   \\ 
			 & \textbf{map} & \textbf{data} & \multicolumn{1}{c}{IIIT5K} & \multicolumn{1}{c}{SVT}  & \multicolumn{1}{c}{IC03} & \multicolumn{1}{c|}{IC13} & \multicolumn{1}{c}{IC15} & \multicolumn{1}{c}{SVTP} & \multicolumn{1}{c}{CT80} \\ 
			\hline
			 CRNN~\cite{CRNN}         & 1D & MJ &$78.2$ & $80.8$ & $-$ & $86.7$ & $-$ & $-$ & $-$ \\
			 RARE~\cite{RARE}         & 1D & MJ &$81.9$ & $81.9$ & $-$ & $-$ & $-$ & $71.8$ & $59.2$ \\
			 STAR-Net~\cite{STAR-Net}    & 1D & MJ+PRI & $83.3$ & $83.6$ & $-$ & $89.1$ & $-$ & $73.5$ & $-$ \\
			 GRCNN~\cite{GRCNN}      &  1D & MJ & $80.8$  & $81.5$ & $-$ & $-$ & $-$ & $-$ & $-$ \\
			 FAN~\cite{FAN}       & 1D & MJ+ST+C & $87.4$ & $85.9$ & $94.2$ & $93.3$ & $-$ & $-$ & $-$ \\
			 ASTER~\cite{Aster}       & STN-1D & MJ+ST & $\textbf{93.4}$ & $\textbf{93.6}$ & $-$ & $91.8$ & $76.1$ & $78.5$ & $79.5$ \\
			 Comb.Best~\cite{Benchmark_Baek}  & STN-1D & MJ+ST & $87.9$  & $87.5$ & $94.4$ & $92.3$ & $71.8$ & $79.2$ & $74.0$ \\
			 ESIR~\cite{ESIR}  & STN-1D & MJ+ST & $93.3$  & $90.2$ & $-$ & $-$ & $76.9$ & $79.6$ & $83.3$ \\
			\hline 
			ATR~\cite{ATR}          & 2D & PRI+C & $-$  & $-$ & $-$ & $-$ & $-$ & $75.8$ & $69.3$ \\
			AON~\cite{AON}      & 2D & MJ+ST & $87.0$ & $82.8$ & $91.5$ & $-$ & $68.2$ & $73.0$ & $76.8$ \\
		    CA-FCN~\cite{2Perspec}  & 2D & ST+C & $92.0$  & $82.1$ & $-$ & $91.4$ & $-$ & $-$ & $79.9$ \\
		     SAR~\cite{SAR}         & 2D & MJ+ST & $91.5$ & $84.5$ & $-$ & $-$ & $69.2$ & $76.4$  & $83.3$ \\
			\hline \hline 
			\model  & 2D & MJ+ST & $92.8$ & $91.3$ & $\textbf{96.7}$ & $\textbf{94.1}$ & $\textbf{79.0}$ & $\textbf{86.5}$ & $\textbf{87.8}$ \\
			\hline
		\end{tabular}
		\caption{\small Scene text recognition accuracies (\%) over seven benchmark test datasets. ``Feature map'' indicates the output shape of image encoder. ``Regular'' datasets consist of horizontally aligned texts and ``irregular'' datasets are made of more diverse text shapes. Accuracies of predicted sequences without dictionary matching are reported. In training data, MJ, ST, C and PRI denote MJSynth, SynthText, Character-labeled, and private data, respectively. 
		}
		\label{table:main_benchmark_result}
	\end{center}
	\vspace{-1.5em}
\end{table*}

We report experimental results on our model, \model. First, we evaluate the accuracy of our model against state of the art methods. We add an analysis on spatial dependencies shown by \model. Second, we assess \model in terms of computational efficiency, namely memory consumption and the number of FLOPs. Third, we conduct ablation studies to evaluate our design choices including the shallow CNN, adaptive 2D positional encoding, and the locality-aware feedforward layer. Finally, we evaluate \model on more challenging cases not covered by current benchmarks, namely rotated and multi-lined texts.

\subsection{STR Benchmark Datasets}
\label{subsec:benchmark}

Seven widely used real-word STR benchmark datasets are used for evaluation~\cite{Benchmark_Baek}. They are divided into two groups, ``Regular'' and ``Irregular'', according to the difficulty and geometric layout of texts.

Below are ``regular'' datasets that contain horizontally aligned texts.
IIIT5K contains 2,000 for training and 3,000 for testing images collected from the web, with mostly horizontal texts. 
Street View Text (SVT) consists of 257 for training and 647 for testing images collected from the Google Street View. Many examples are severely corrupted by noise and blur. 
ICDAR2003 (IC03) contains 867 cropped text images taken in a mall.
ICDAR2013 (IC13) consists of 1015 images inheriting most images from IC03. 

``Irregular'' benchmarks contain more texts of arbitrary shapes. 
ICDAR2015 (IC15) contains 2077 examples more irregular than do IC03 and IC13. Street View Text Perspective (SVTP) consists of 645 images which text are  typically captured in perspective views. 
CUTE80 (CT80) includes 288 heavily curved text images with high resolution. Samples are taken from the real world scenes in diverse domains.

\subsection{Implementation Details}
\label{subsec:Imple}

\subsubsection{Training set} 
Two widely used training datasets for STR are Mjsynth and SynthText. Mjsynth is a 9-million synthetic dataset for text recognition, generated by Jaderberg~\etal~\cite{MJSynth}. SynthText represents 8-million text boxes from 800K synthetic scene images, provided by Gupta~\etal~\cite{SynthText}. Most previous works have used these two synthetic datasets to learn diverse styles of synthetic sets, each generated with different engines. \model is trained on the combined training set, SynthText$+$Mjsynth, as suggested in Baek~\etal~\cite{Benchmark_Baek} for fair comparison.

\subsubsection{Architecture details} 
Input images are resized to $32\times100$ both during training and testing following common practice. 
The number of hidden units for self-attention layers is 512, and the number of filter units for feedforward layers is 4-times of the hidden unit. The number of self-attention layers in encoder and decoder are $N_e=12$ and $N_d=6$. The final output is a vector of 94 scores; 10 for digits, 52 for alphabets, 31 for special characters, and 1 for the end token. 

\subsubsection{Optimization}
Our model has been trained in an end-to-end manner using the cross-entropy loss. We have applied image rotation augmentation, where the amount of rotation follows the normal distribution $N(0,(34\degree )^2)$. 
\model is trained with Adam optimizer \cite{Adam} with the initial learning rate 3e-4. Cyclic learning rate~\cite{clr} has been used, where the cycle step is 250,000. Batch size is 256, and the learning is finished after 4 epochs.
In our ablation study, we applied this optimization method on our baseline models for fair comparison.

\subsubsection{Evaluation}
We trained our model with spacial characters, adopting the suggestion by~\cite{Benchmark_Baek}. When we evaluate our model, we calculate the case-insensitive word accuracy~\cite{Aster}. Such training and evaluation method has been conducted in recent STR papers~\cite{Aster,SAR,2Perspec}. In our ablation studies, we use the unified evaluation dataset of all benchmarks (8,539 images in total) as done in~\cite{Benchmark_Baek}.

\subsection{Comparison against Prior STR Methods}
\label{subsec:results_on_benchmarks}

We compare the \model performance against existing STR models in Table~\ref{table:main_benchmark_result}. The accuracies for previous models are reported accuracies. Methods are grouped according to the dimensionality of feature maps, and whether the spatial transformer network (STN) has been used.
The STN module and 2D feature maps have been designed to help recognizing texts of arbitrary shapes. 
We observe that \model outperforms other 2D approaches on all benchmarks and that it attains the best performance on five of them against all prior methods considered. In particular, on irregular benchmarks that we aim to solve, \model improves upon the second best method with a large margin of 4.7 pp on average.

\subsection{Comparing \model against SAR}

\begin{table}[b]
\tabcolsep=0.15cm
\small
\centering
\renewcommand{\arraystretch}{1.2}
\begin{tabular}{cc|c|c|c} 
 \hline
 \textbf{Encoder} & \textbf{Decoder} & \textbf{Params} &\textbf{FLOPs} & \textbf{Accuracy} \\
  \hline
 ResNet(2D) & LSTM  & 56M & 21.9B & 87.9 \\ 
 \model(2D) & LSTM  & 44M & 16.4B & 88.9 \\ 
 ResNet(2D) & \model  & 67M & 41.4B & 88.3\\ 
\model(2D) & \model  & 55M & 35.9B &  89.2 \\ 
 \hline
\end{tabular}
\caption{\small Impact on accuracy and efficiency (the number of parameters and FLOPs) incurred by \model encoder and decoder. The first row corresponds to SAR~\cite{SAR} and the last is the proposed \model (ours).}
\label{table:effect_of_selfatt2}
\vspace{-0.5em}
\end{table}

Since \model shares many similarities with SAR~\cite{SAR}, where the difference is the choice of encoder (self-attention versus convolutions) and decoder (self-attention versus LSTM), we provide a more detailed analysis through a thorough comparison against SAR. We analyze the accuracy-efficiency trade-off as well as their qualitative differences.

\subsubsection{Accuracy-efficiency trade-off}
We analyze the contributions of self-attention layers in \model encoder and decoder, focusing both on the accuracy and efficiency. See Table~\ref{table:effect_of_selfatt2} for ablative analysis. The baseline model is SAR~\cite{SAR} given in the first row (ResNet encoder with 2D attention LSTM decoder), and one can partially update SAR by replacing either only the encoder or the decoder of \model. 

We observe that upgrading ResNet encoder to \model encoder improve the accuracy by 1.0 pp and 0.9 pp over LSTM and \model decoders, respectively, while actually improving the space and time efficiency (reduction of 12M parameters and 5.5B FLOPs in both cases). This is the result of inherent computational efficiency enjoyed by self-attention layers and careful design of \model encoder to reduce FLOPs by modeling long-term and short-term dependencies of the features efficiently. The \model decoder, which is nearly identical to the original Transformer decoder, does provide further gain of 0.3 pp accuracy boost, but at the cost of increased memory consumption (+11M) and FLOPs (+19.5B).

\begin{figure}[t]
    \centering
    \includegraphics[width=1.0\linewidth]{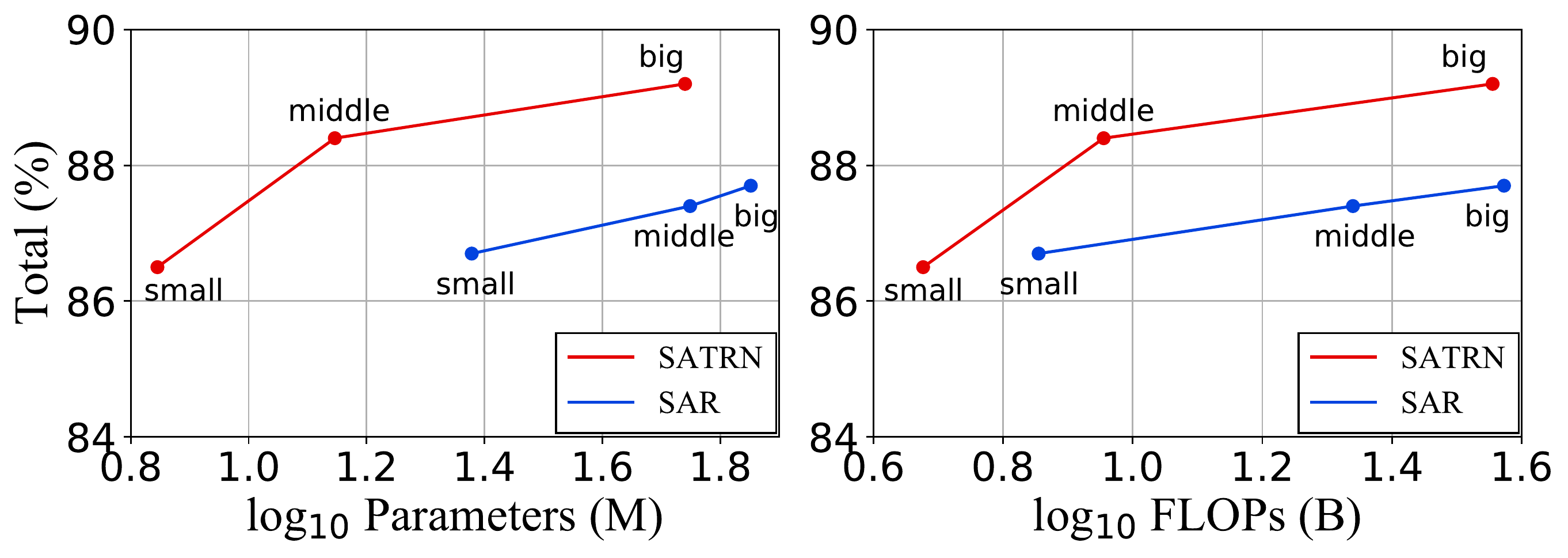}
    \caption{\small
    Accuracy-efficiency trade-off plots for SAR and \model. We have made variations, small, middle, and big, to control over the number of layers.}
    \label{fig:comp_w_sar}
    \vspace{-1em}
\end{figure}

To provide a broader view on the computational efficiency due to self-attention layers, we have made variations over SAR~\cite{SAR} and \model with varying number of layers. 
The original SAR contains ResNet34 as an encoder (SAR-middle), and we consider replacing the encoder with ResNet18 (SAR-small) and ResNet101 (SAR-big). Our base construct \model is considered \model-big. We consider reducing the channel dimensions in all layers from 512 to 256 (\model-middle) and further reducing the number of encoder layers $N_e=9$ and that of decoder layers $N_d=3$ (\model-small). 

Figure \ref{fig:comp_w_sar} compares the accuracy-cost trade-offs of SAR~\cite{SAR} and \model. We observe more clearly that \model design involving self-attention layers provides a better accuracy-efficiency trade-off than SAR approach. We conclude that for addressing STR problems, \model design is a favorable choice.

\begin{figure}[t]
    \centering
    \subfloat[\small Character ROI]
    {\includegraphics[width=0.31\linewidth]{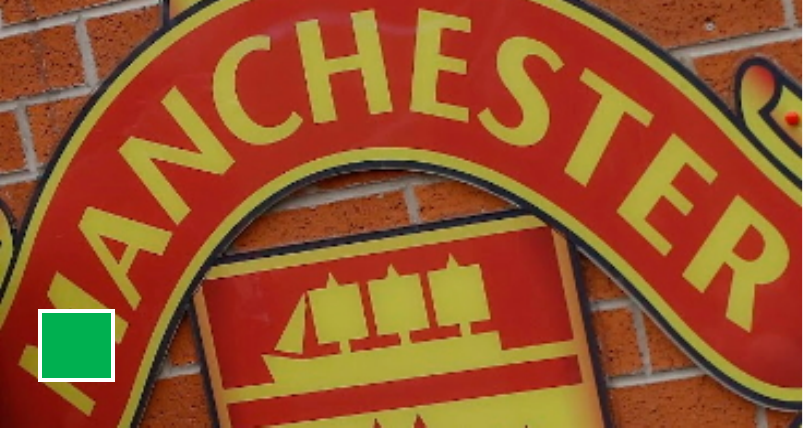}} \hfill
    \subfloat[\small SA at depth 1] {\includegraphics[width=0.31\linewidth]{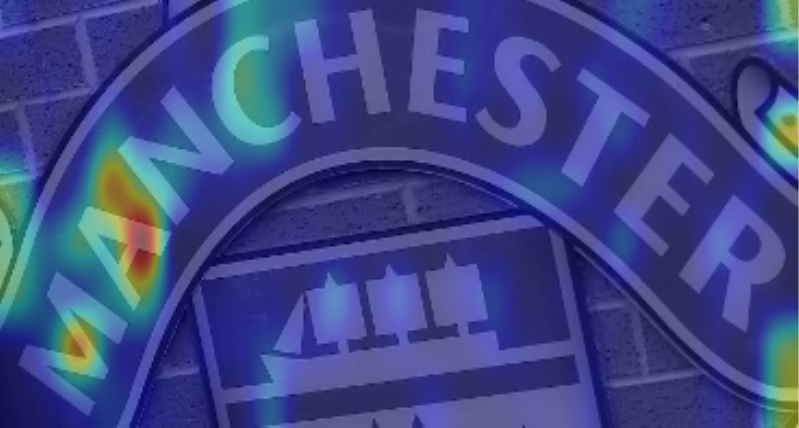}} \hfill
    \subfloat[\small SA at depth 2] {\includegraphics[width=0.31\linewidth]{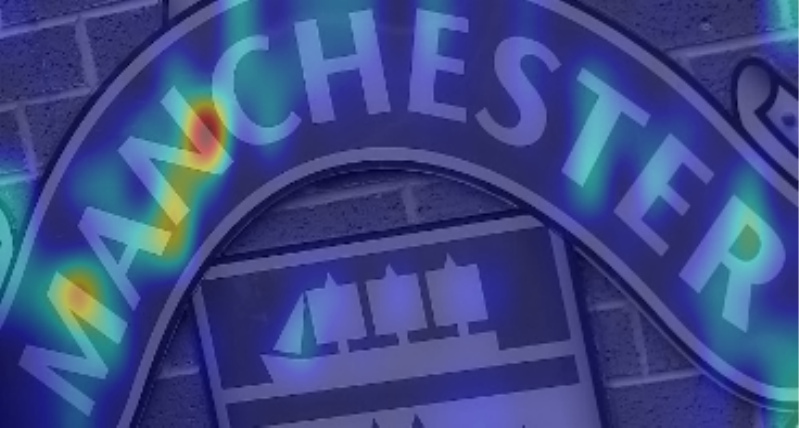}}
    \caption{\small Visualization of the self-attention maps. See text.
    }
    \label{fig:self-attn}
    \vspace{-1em}
\end{figure}

\subsubsection{Qualitative comparison}
We provide a qualitative analysis of how the 2D self-attention layers in encoder extract informative features. Figure \ref{fig:self-attn} show the human-defined character region of interest (ROI) as well as the corresponding self-attention heatmaps (SA) at depth $n$, generated by propagating the character ROI from the last layer to $n$ layers below through self-attention weights. It shows the supporting signals relations at $n$ for recognizing the designated character. 

We observe that for character `M' the last self-attention layer identifies the dependencies with the next character `A'. SA at depth 2 already propagates the supporting signal globally, taking advantage of long-range connections in self-attention. By allowing long-range computations within small number of layers, \model achieves a good performance while removing redundancies created by accumulating local information too many times (convolutional encoder).

\subsection{Ablation Studies on Proposed Modules}

\model encoder is made of many design choices to adapt Transformer to the STR task. We report ablative studies on those factors in the following part, and experimentally analyze alternative design choices. The default model used hereafter is \model-small.

\subsubsection{Adaptive 2D positional encoding (A2DPE)}

\begin{table}
\small
\tabcolsep=0.05cm
\centering
\subfloat[Positional encoding\label{table:effect_of_pe}]
{\begin{tabular}{l|c} 
 \hline
 \textbf{Positional} & \multirow{2}{*}{\textbf{Accuracy}} \\
 \textbf{encoding}\\
 \hline
SATRN-small &  \\
\quad + None & 83.8\\ 
\quad + 1D-Flat & 85.8\\ 
\quad + 2D-Concat & 85.8 \\ 
\quad + A2DPE & \textbf{86.5}\\ 
 \hline
\end{tabular}
}
\hfill
\subfloat[Downsampling\label{table:effect_of_size}]
{
\begin{tabular}{cc|c|c} 
 \hline
 \multicolumn{2}{c|}{\textbf{Downsampling}} & \multirow{2}{*}{\textbf{FLOPs}}  & \multirow{2}{*}{\textbf{Accuracy}} \\
 height & width &  & \\
 \hline
 1/2 & 1/2 & 8.4B &  86.9 \\ 
 1/4 & 1/4  & 4.7B &  86.5 \\ 
 1/8 & 1/4 & 2.4B &  85.5 \\ 
 1/16 & 1/4 & 1.3B &  83.6 \\ 
 1/32 (1D) & 1/4 & 0.7B &  81.9 \\ 
 \hline
\end{tabular}
}
\caption{Performance of \model-small with different positional encoding (PE) schemes and downsampling rates.}
\vspace{-0.7em}
\end{table}

\begin{figure}
    \vspace{-1em}
    \centering
    \subfloat[ \small $r\in(0,0.6)$  \label{fig:low_pe}] {\includegraphics[width=0.3\linewidth]{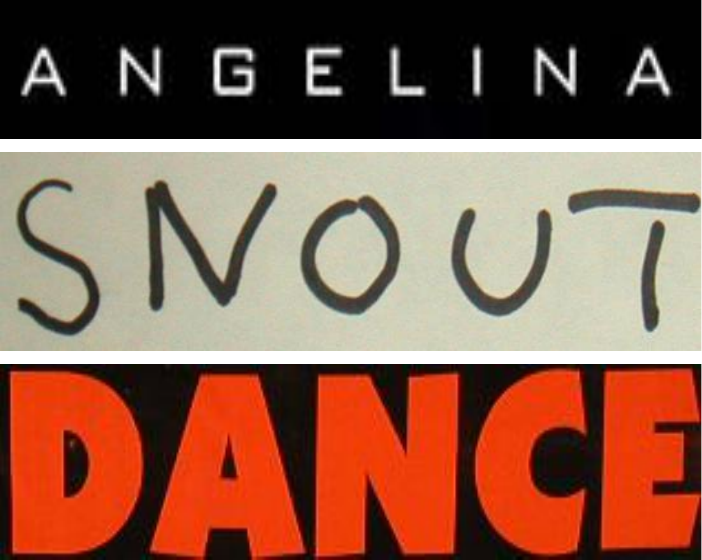}} \hfill
    \centering
    \subfloat[ \small $r\in(0.6,0.8)$ \label{fig:mid_pe}] {\includegraphics[width=0.3\linewidth]{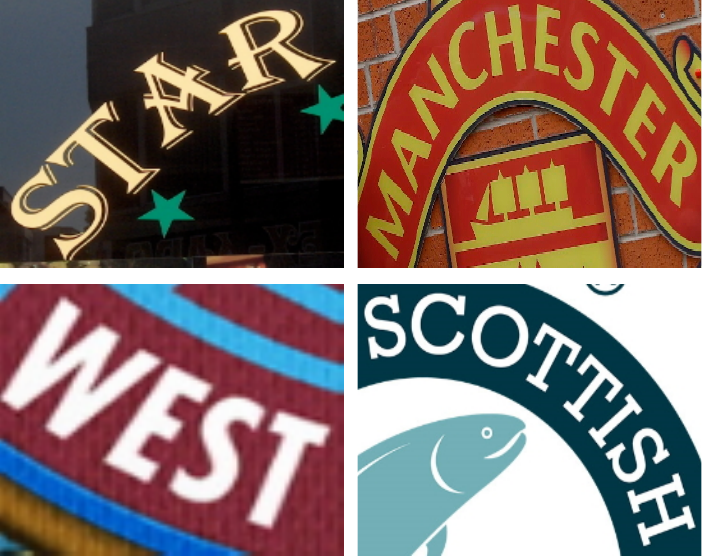}} \hfill
    \centering
    \subfloat[ \small $r\in(0.8,\infty)$]
    {\includegraphics[width=0.3\linewidth]{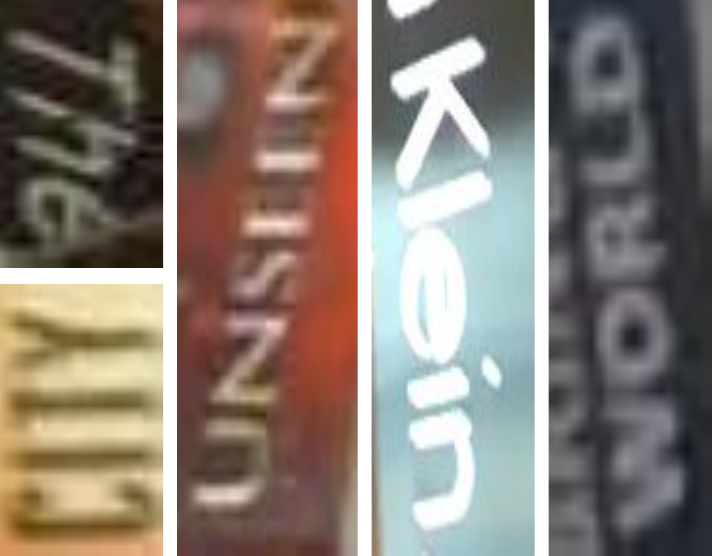}}
    \caption{\small Examples in three groups separated by the range of the aspect ratios, $ r=|| \boldsymbol{\alpha}||_1 / ||\boldsymbol{\beta}||_1 $.} 
    \label{fig:pe_ratio}
\end{figure}

This new positional encoding is necessary for dynamically adapting to the inherent aspect ratios incurred by overall text alignment (horizontal, diagonal, or vertical).  
As alternative options, we consider not doing any positional encoding at all (``None'')~\cite{SAGAN,nonlocal}, using 1D positional encoding over flattened feature map (``1D-Flatten''), using concatenation of height and width positional encodings (``2D-Concat'')~\cite{Image-Transformer}, and the A2DPE that we propose. See Table \ref{table:effect_of_pe} for the results. We observe that A2DPE provides the best accuracy among four options considered. 

We visualize random input images from three groups with different predicted aspect ratios, as a by-product of A2DPE. Figure \ref{fig:pe_ratio} shows the examples according to the ratios $||\boldsymbol{\alpha}||_1/||\boldsymbol{\beta}||_1$. Low aspect ratio group, as expected, contains mostly horizontal samples, and high aspect ratio group contains mostly vertical samples. By dynamically adjusting the grid spacing, A2DPE reduces the representation burden for the other modules, leading to performance boost.

\subsubsection{Locality-aware feedforward layer}

\begin{figure}[t]
    \centering
    \small
    \includegraphics[width=1.0\linewidth]{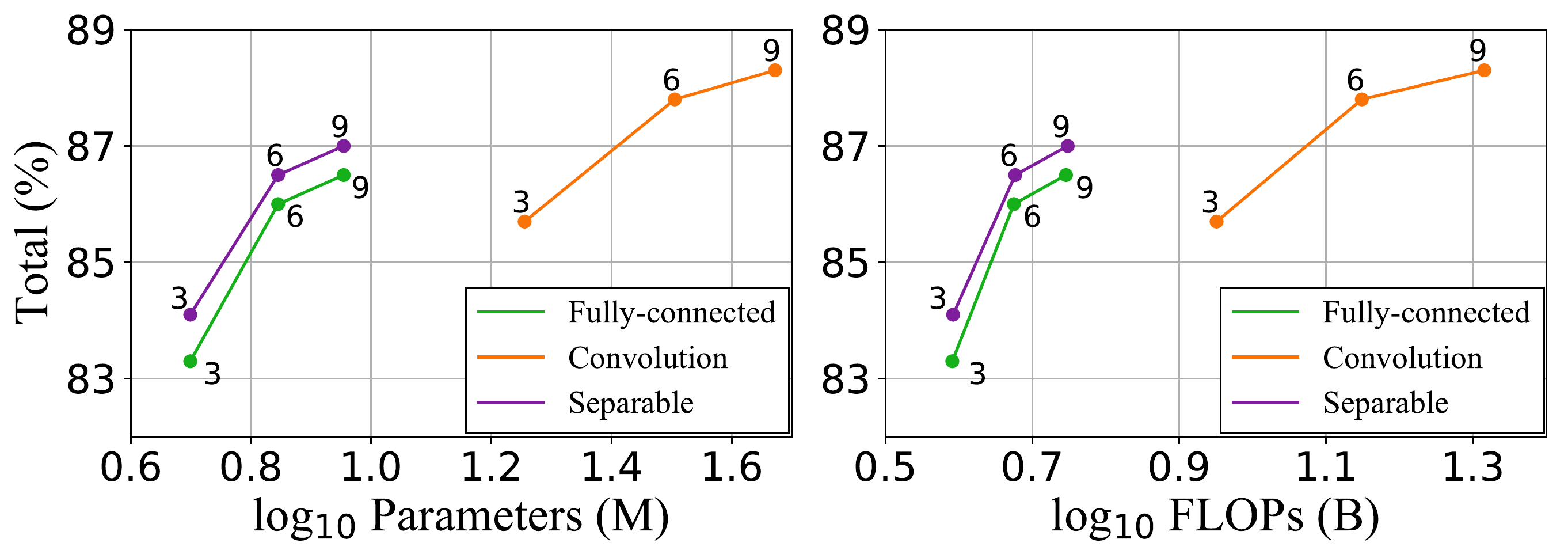}
    \caption{\small Performance comparison of the feedforward blocks according to the number of parameters and FLOPs. Numbers above data points denote the number of encoding layers.}
    \label{fig:ff_blocks}
    \vspace{-1em}
\end{figure}

We have replaced the point-wise feedforward layers in the Transformer encoder~\cite{Transformer} with our novel locality-aware feedforward layers to seek performance boost at low extra cost. To analyze their effects, we consider the two alternatives described in Figure~\ref{fig:ff}, each with different number of encoder layers (3, 6, or 9).

The resulting accuracy-performance trade-offs are visualized in Figure~\ref{fig:ff_blocks}. Compared to the point-wise feedforward, naive convolution results in improved accuracy, but roughly with four times more parameters and FLOPs. We alleviate the computation cost with depth-wise convolutions (locality-aware feedforward) and achieve a better accuracy at nearly identical computational costs. 

\subsubsection{Feature map height}

Finally, we study the impact of the spatial degree of freedom in the 2D feature map of \model-small on its accuracy and computational costs. We interpolate between \model-small using full 2D feature map and the same model using 1D feature map by controlling the downsampling rate along height and width dimensions. \model-small is using the 1/4 downsampling factor for both height and width, and we consider further downsampling the height sequentially with 1/2 factor, until only 1 height dimension is left (1/32 height downsampling). To see the other extreme, we have considered downsampling less (downsample only 1/2 for both width and height).

Table \ref{table:effect_of_size} shows the results. There is a consistent drop in FLOPs and accuracy as the feature map sizes are reduced. When height is downsampled with rate greater than 1/8, performances drop dramatically (more than 2.9 pp). The results re-emphasize the importance of maintaining the 2D feature maps throughout the computation.

\subsection{More Challenges: Rotated and Multi-Line Text}

Irregular text recognition benchmarks (IC15, SVTP, and CT80) are attempts to shift the focus of STR research to more difficult challenges yet to be address by the field. While these datasets do contain texts of more difficult shapes, it is not easy to analyze the impact of the \textit{type} and \textit{amount} of shape distortions. We have thus prepared new synthetic test sets (transformed from IC13) that consists purely of single type and degree of perturbation. Specifically, we measure the performance against texts with varying degrees of rotations ($0\degree$, $90\degree$, $180\degree$, and $270\degree$) as well as multi-line texts.

We compare against two representative baseline models,  FAN~\cite{FAN} and SAR~\cite{SAR}. Optimization and pre-processing details including training dataset and augmentation are unified for fair comparison.

\begin{table}[t!]
\tabcolsep=0.2cm
\centering
\small
\begin{tabular}{c|cccc|c} 
 \hline
 \multirow{2}{*}{\textbf{Model}} & \multicolumn{4}{c|}{\textbf{Rotated (IC13)}} & \multirow{2}{*}{\textbf{Multi-line}} \\
  & $0\degree$ & $90\degree$ & $180\degree$ & $270\degree$ &  \\ [0.5ex]
 \hline
  FAN (1D) & 87.0 & 81.9 & 86.8 & 84.1 & 44.7 \\
  SAR (2D) & 88.5 & 88.4 & 89.1 & 88.8 & 46.7 \\ 
  \model (2D) & \textbf{90.7} & \textbf{90.5} & \textbf{91.6} & \textbf{91.5}& \textbf{63.8} \\
 \hline
\end{tabular}
\caption{\small The results on two challenging text datasets; heavily rotated text and mutli-line text.}
\label{table:irregular_text_table}
\vspace{-1em}
\end{table}

\subsubsection{Rotated text} 
Most STR models based upon the horizontal text assumption cannot handle heavily rotated texts. \model on the other hand does not rely on any such inductive bias; its ability to recognize rotated texts purely depends upon the ratio of such cases shown during training. To empirically validate this, we have trained the models with wider range of rotations: $\text{Uniform}(0\degree, 360\degree)$. Input images are then resized to $64\times64$. Second column group in Table \ref{table:irregular_text_table} shows the results of rotated text experiments.
We confirm that \model outperforms FAN and SAR while retaining stable performances for all rotation levels.

\subsubsection{Multi-line text}
We analyze the capability of models on recognizing multi-line texts, which would require the functionality to change line during inference.
We have synthesized multi-line texts using SynthText and MJSynth for training the models. For evaluation we have utilized multi-line text manually cropped from the scene images in IC13. 
Last column in Table \ref{table:irregular_text_table} shows the results.
\model indeed performs better than the baselines, showing its capability to make a long-range jump to change line during inference.

\begin{figure}[h]
    \centering
    \includegraphics[width=1.0\linewidth]{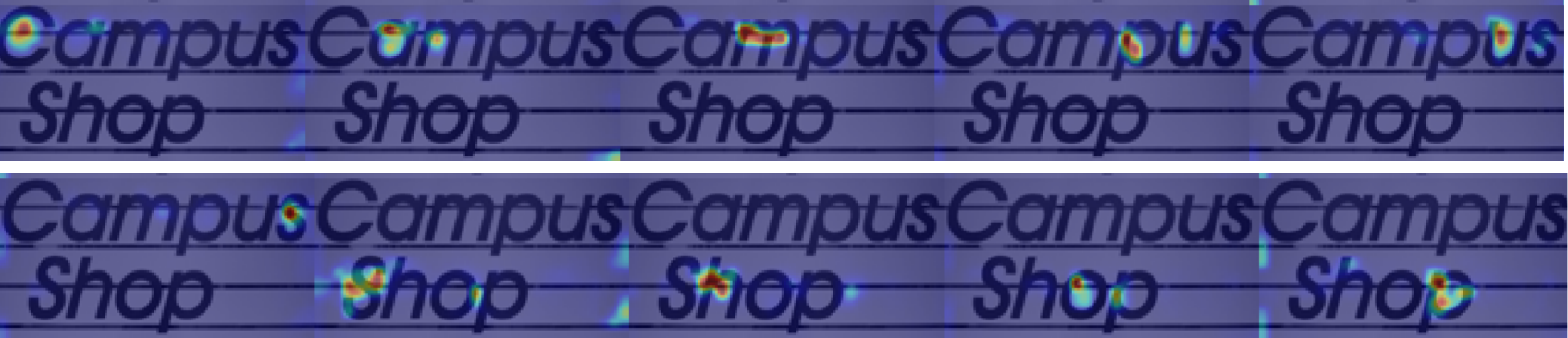}
    \caption{The 2D attention maps on a multi-line example. The 2D attention follows the first text line and then moves to the next line.}
    \label{fig:rotation_graph}
    \vspace{-1em}
\end{figure}

Figure \ref{fig:rotation_graph} shows the attention map of the \model decoder to  retrieve 2D features. \model distinguishes the two lines and successes to track the next line. The results show that \model enables the 2D attention transition from the current region to a non-adjacent region on the image. 

\section{Conclusions}

Scene text recognition (STR) field has seen great advances in the last couple of years. Models are now working well on texts of canonical shapes. We argue that the important remaining challenge for STR is the recognition of texts with arbitrary shapes. To address this problem, we have proposed the \textit{Self-Attention Text Recognition Network (\model)}. By allowing long-range dependencies through self-attention layer, \model is able to sequentially locate next characters even if they do not follow canonical arrangements. We have made several novel modifications on the Transformer architecture to adapt it to STR task. We have achieved the new state of the art performances on irregular text recognition benchmarks with great margin (5.7 pp boost on average). \model has shown particularly good performance on our more controlled experiments on rotated and multi-line texts, ones that constitute the future STR challenges. We will open source the code.

\bibliography{references}
\bibliographystyle{aaai}
\end{document}